\documentclass{article}

     \PassOptionsToPackage{numbers, compress}{natbib}

     \usepackage[final]{neurips_2020}

\usepackage[utf8]{inputenc} 
\usepackage[T1]{fontenc}    
\usepackage{hyperref}       
\usepackage{url}            
\usepackage{booktabs}       
\usepackage{amsfonts}       
\usepackage{nicefrac}       
\usepackage{microtype}      

\usepackage{soul}
\usepackage{color}

\newcommand{\etal}{\textit{et al}. }
\newcommand{\ie}{\textit{i}.\textit{e}. }
\newcommand{\eg}{\textit{e}.\textit{g}. }

\usepackage{multirow}
\usepackage{MnSymbol}
\DeclareMathOperator*{\argmax}{argmax}

\DeclareMathOperator*{\mode}{mode}

\title{Performance Variability in Zero-Shot Classification}

\author{
   Matías Molina\\
   Universidad Nacional de C\'ordoba \\
  \texttt{matias.molina@unc.edu.ar} \\
   \And
   Jorge S\'anchez \\
   CONICET \\
   \texttt{jorge.sanchez@unc.edu.ar} \\
}
\begin{document}

\maketitle

\begin{abstract}
  Zero-shot classification (ZSC) is the task of learning predictors for classes
  not seen during training. Although the different methods in the literature
  are evaluated using the same class splits, little is known about their
  stability under different class partitions. In this work we show
  experimentally that ZSC performance exhibits strong variability under
  changing training setups. We propose the use ensemble learning as an attempt
  to mitigate this phenomena.
\end{abstract}

\section{Motivation}

Training classifiers on specific non-generic domains requires a non negligible
effort on data annotation. Although this might be easy for some of the target
categories, it might become too costly for others due to the long-tail
distribution of samples. This has motivated the development of models that can
be trained on little data (few-shot learning) or no data at all (zero-shot
learning). In zero-shot classification (ZSC) \cite{lampert2009} we are given a
set of labeled samples from a known set of categories and the goal is to learn
a model that is able to cast predictions over a set of categories not seen
during training. Despite having identified the difficulties and particularities
in the evaluation of different approaches \cite{xian2018zero}, little attention
has been paid to the effect of considering different training class partitions
for a given problem. Given the large number and diversity of models proposed in
the literature in the recent years, we believe this is an important factor to
be considered when choosing between competing approaches. In this work, we
start exploring this problem. Our preliminary experiments using different
datasets of varying granularity and two simple baselines confirm our
hypothesis: performance differences observed in the literature might be not as
significant as it seems due to the large variability observed across different
subsets of training classes.

\section{Experiments and Discussion}

In ZSC we are given a training set
$\mathcal{D}^{tr} = \{(x_i, y_i)~|~ x_i \in \mathcal{X}, y_i \in
\mathcal{Y}^{tr}\}$.  The goal is to learn a mapping
$f:\mathcal{X}\rightarrow\mathcal{Y}$ from $\mathcal{D}^{tr}$ that can be used
to classify samples over a different set $\mathcal{Y}^{ts}\subset\mathcal{Y}$.
We consider the standard ZSC setting, where
$\mathcal{Y}^{tr} \cap \mathcal{Y}^{ts} = \emptyset$.  Given a representation
$z_y \in \mathbb{R}^E$ for each $y \in \mathcal{Y}$, a common approach
\cite{akata-15, eszsl} is to learn a function
$F:\mathcal{X}\times\mathcal{Z}\rightarrow\mathbb{R}$ to reflect the degree at
which $x$ and $z$ agree on a given concept. Given a test sample $x$, its class
is predicted as $\hat{y} = \argmax_{y\in \mathcal{Y}^{ts}} F(x, z_y)$.

The work of Xian \etal \cite{xian2018zero} identified several problems in the
evaluation methodology used in the ZSC literature. One key contribution of
their work was the proposal of fixed set of train/test class splits for
different datasets. Although this addresses many of the evaluation problems
identified in \cite{xian2018zero}, it does not considers the effect of varying
training class partitions. We believe analyzing not only the mean but also the
variability of the zero-shot predictive performance under changing training
configurations is an important factor towards a more thoughtful evaluation of
the different methods. Our work is a first step in that direction.

Table~\ref{table:sje_eszsl_pvalue} shows the mean and standard deviation over
different training partitions for two simple baselines, ESZSL\cite{eszsl} and
SJE\cite{akata-15}, on two fine-grained (SUN\cite{sun} and CUB\cite{cub}) and
two coarse-grained (AWA1\cite{lampert2009} and AWA2\cite{xian2018zero})
datasets, using different class partitions sampled at random. We observe a
great deal of variability whether the sample per class imbalance is considered
(avg. acc.)  or not (avg. per-class acc.). We observe that the difference in
performance (as reported in the literature) might bias the selection of one
method over the other even when their difference is not statistically
significant. The table also show p-values of a Wilcoxon signed-rank test
computed from 22 different partitions chosen at random. We see that while for
the fine-grained cases we can reject the null hypothesis for a fairly low
confidence level, this is not the case in the coarse-grained data regime.
Although the difference in mean values seems high (for the standards observed
in the literature), the variability observed in the experiments warns against
choosing one method over the other.

\paragraph{Ensemble learning for ZSC.}

Beyond the identification of the variability problem, we ran experiments using
standard ensemble techniques as an attempt to mitigate its effect. The idea is
that by combining more than one predictor into a single model, it is possible
to reduce the variance by averaging \cite{dietterich2000ensemble}. One popular
approach is the \textit{Bootstrap Aggregation} or \textit{Bagging}
meta-algorithm \cite{breiman1996bagging}. It is based on learning different
predictors using different subsets of training samples and aggregating them via
a suitable voting scheme. The \emph{hard} voting scheme assigns the class
predicted by the majority, \ie $\hat{f}(x) = \mode\{f_1(x),...,f_n(x)\}$. In
\emph{soft} voting, prediction is given by the highest score over all the
models , \ie $\hat{f}(x) = \argmax_y \{\sum_i F_i(x, z_y)\}$.

In the context of ZSC, we use different (random) subsets of training categories
to generate the set of base predictors, \ie we learn a set $n$ predictors using
a proportion $s$ of randomly chosen classes from the original training set. We
use hard and soft ensembles considering $n = \{ 10, 30, 50, 70, 90\}$ and
$s = \{0.3, 0.5, 0.7, 0.9\}$, \eg $(n,s)=(10,0.3)$ means training $10$
different models using $30\%$ of the full set of training categories. Each
sub-problem is trained on a different subset of training classes. We sample 4
different sub-problems for each $(n,s)$ combination. Baseline performances are
as follows: $56.91~(1.63)$ on SUN, $54.80~(2.82)$ on CUB, and $70.62~(7.32)$,
$73.26~(4.81)$ on AWA1 and AWA2 respectively.  We use the ResNet101 features
and continuous attribute vectors from \cite{xian2018zero} and normalize both to
unit norm. We found that as the proportion $s$ increases performance approaches
the baseline, which is to be expected since the set of training categories
tends to resemble the original set. The standard deviation may marginally
decrease but with a considerable loss in performance. This situation is more
noticeable in the case of AWA1 and AWA2, both coarse-grained datasets, compared
to the others. Table~\ref{table:robustness_apc_soft} shows the ensemble results
for $n=90$. \footnote{Different combinations of voting schemes and accuracy
  metrics lead to similar conclusions.}  Beyond these observations, the use of
ensemble does not lead to an increase on the overall ZSC
performance. Alternatives to this formulation is the topic of our current
research.

\begin{table}
  \centering
  \caption{Soft ensemble results. Top-1 average per-class accuracy and its std. deviation, for $n=90$.} 
  \label{table:robustness_apc_soft}
\begin{tabular}{l|lllll}
\toprule
	& $s$ & 0.3          & 0.5          & 0.7          & 0.9          \\
\midrule
SUN & &  55.61   (2.16) &  56.81   (2.02) &  56.77   (1.98) &  57.03  (1.73) \\
CUB
 & &  50.89   (2.92) &  53.45   (2.84) &  54.39   (2.84) &  54.83   (2.72) \\
AWA1
 & & 65.35   (6.52) &  68.38   (7.49) &  69.70   (7.63) &  70.52   (7.31) \\
AWA2
 & &  66.90    (3.70) &  70.39   (4.23) &  72.16   (4.26) &  73.13  (4.52) \\
\bottomrule
\end{tabular}
\end{table}

\begin{table}
\caption{Top-1 accuracy and average top-1 per-class accuracy, std. deviation and p-value for 22 splits.}
\label{table:sje_eszsl_pvalue}
\begin{tabular}{llllll}
\toprule
                                {}   &   {}       & SUN          & CUB          & AWA1         & AWA2 \\
\midrule
\multirow{3}{*}{Avg. acc.}           &   ESZSL    & 55.90 (1.95) & 53.49 (2.10) & 69.66 (9.94) & 71.10 (10.94) \\
				     &   SJE      & 59.16 (2.37) & 56.08 (3.03) & 68.85 (7.96) & 68.84 (11.16) \\
				     &   p-value  & 0.000001     & 0.0012       & 0.7024       & 0.5028 \\
\midrule
\multirow{3}{*}{Avg. per-class  acc.} & ESZSL    & 55.92 (1.94) & 53.81 (2.20) & 69.34 (9.02) & 71.48 (9.54) \\
				       &  SJE     & 59.73 (2.17) & 56.19 (2.44) & 69.48 (8.27) & 69.34 (9.63) \\
				       & p-value  & 0.0000005    & 0.0000024    & 0.8736       & 0.1762 \\
\bottomrule
\end{tabular}
\end{table}

\newpage

\medskip
\small

\end{document}